\documentclass[10pt,letterpaper]{article}

\usepackage{cogsci}

\cogscifinalcopy 

\usepackage{hyperref}
\usepackage{graphicx}
\usepackage{pslatex}
\usepackage{apacite}
\usepackage{array}
\usepackage{multirow}
\usepackage{cellspace}

\usepackage{float} 
\usepackage{color}
\usepackage{xcolor}
\usepackage{natbib}

\DeclareTextFontCommand\textcourier{\fontfamily{qcr}}
\newcommand{\figref}[1]{Fig.~\ref{#1}}

\title{Analyzing the Roles of Language and Vision in Learning from Limited Data}

\author{
{\large \bf Allison Chen$^1$, Ilia Sucholutsky$^1$, Olga Russakovsky$^1$, Thomas L. Griffiths$^{1,2}$} \\
{$^1$ Department of Computer Science, Princeton University} \\
{$^2$ Department of Psychology, Princeton University} \\
\{\texttt{allisonchen, is2961, olgarus, tomg\}@princeton.edu} \\
}

\begin{document}

\maketitle

\begin{abstract}
Does language help make sense of the visual world? How important is it to actually see the world rather than having it described with words? These basic questions about the nature of intelligence have been difficult to answer because we only had one example of an intelligent system -- humans -- and limited access to cases that isolated language or vision. However, the development of sophisticated Vision-Language Models (\textbf{V}LMs) by artificial intelligence researchers offers us new opportunities to explore the contributions that language and vision make to learning about the world. We ablate components from the cognitive architecture of these models to identify their contributions to learning new tasks from limited data. We find that a language model leveraging all components recovers a majority of a \textbf{V}LM's performance, despite its lack of visual input, and that language seems to allow this by providing access to prior knowledge and reasoning.

\end{abstract}

\section{Introduction}

Language is effective in communicating visual ideas between two individuals, but what is its role in \textit{enabling} visual understanding and intelligence? Perhaps we might attribute the effectiveness of language in conveying visual ideas to its combination with visual input through reasoning. But is it possible that language itself can facilitate visual understanding? If so, to what extent? Intuitively, we might think that seeing the world is necessary for understanding its visual structure, but studies show that congenitally blind (i.e., blind since birth) individuals have mental representations of color similar to those of sighted individuals \citep{kim2021shared}. Similar results have been found for other visual tasks involving spatial memorization, shape understanding, and visual concept association \citep{kerr1983role, aleman2001visual, zimler1983imagery, connolly2007effect}.

Until now, studies like these that investigate where nature has isolated language or vision have offered our best chance at identifying the roles that these different capacities have in our understanding of the world around us. However, recent developments in artificial intelligence offer another way to tease apart the contributions of language and vision. They provide tools to create new kinds of intelligent systems, enabling us to explore the implications of using cognitive architectures that differ in inputs and components.

Multi-modal Vision-Language Models (\textbf{V}LMs) \citep{radford2021learning, liu2023llava, singh2022flava} are ideal examples of systems that can be used to engage with these classic questions. \textbf{V}LMs are built by combining vision models with Large Language Models (LLMs) and have demonstrated impressive visual recognition and understanding \citep{sun2022dualcoop, guo2023texts, rao2022denseclip}. By studying a \textbf{V}LM and its corresponding LLM backbone, we can isolate vision and language capabilities to analyze their respective contribution to visual understanding. By simulating systems with different components, we can explore how the cognitive architectures of these systems affect their performance. 

The key contribution of our work is understanding how each component in a \textbf{V}LM, especially the language-related ones, contributes to its performance. We focus on three internal components: visual processing, prior knowledge, and reasoning. We start with a \textbf{V}LM, break down its cognitive architecture using related models, and analyze each component's contribution to performance on learning new tasks from limited data. Our results show that removing visual processing but retaining full language capabilities maintains just over 75\% of the \textbf{V}LM performance. Furthermore, when any one of the language components is missing, performance drops significantly, indicating each one is necessary. Conversely, reducing a \textbf{V}LM to a vision-only model cuts performance in half and is comparable to an incomplete language model. These results suggest that language may indeed play a key role in understanding the visual world, allowing us to use prior knowledge and reasoning to make sense of our sensory information.

\section{Background}
In this section, we will first review LLMs and \textbf{V}LMs then summarize results in cognitive science that support the idea that language may enable visual understanding.

\subsection{Large Language Models \& Vision-Language Models}

LLMs are large computational models, typically possessing billions of parameters, that use massive amounts of text data to optimize the next word prediction in a given sequence. Common models include the GPT family \citep{radford2018improving, radford2019language, brown2020language, openai2023gpt4}, LLaMA \citep{touvron2023llama}, and Vicuna \citep{zheng2023judging}. These models use the Transformer architecture \citep{vaswani2017attention} to learn which previous words to attend to when predicting the next word. Much of an LLM's cognitive abilities can be attributed to one of two components: reasoning and prior knowledge. Reasoning provides the ability to adapt to new tasks via in-context learning, i.e., providing examples in the prompt \citep{brown2020language, gao2020making, bansal2020self}. Prior knowledge enables LLMs to answer factual questions \citep{petroni2019language, heinzerling2020language}. In our work, we aim to isolate these two components by varying the task at hand to understand how they each contribute to performing visual recognition.

\textbf{V}LMs connect LLMs to vision models. Early \textbf{V}LMs such as CLIP \citep{radford2021learning}, ALIGN \citep{jia2021scaling}, and DeCLIP \citep{li2021supervision} aligned the representation spaces between text and image encoders using methods such as contrastive learning \citep{chopra2005learning, hadsell2006dimensionality}. However, as models grew in size, training them became computationally prohibitive. As a result, newer \textbf{V}LMs such as LLaVA \citep{liu2023llava} adapt pre-trained image encoders \citep{dosovitskiy2020image, radford2021learning} to fit the representation space of a pre-trained LLM. \textbf{V}LMs have been applied to video understanding \citep{wu2023zero, zhao2023antgpt}, robotic planning \citep{ahn2022can, driess2023palm}, and even geometry-based tasks such as pose estimation \citep{feng2023posegpt}. This widespread adoption suggests that language enhances a vision model's capabilities. However, by entangling the two, it is difficult to pinpoint what language specifically contributes to visual understanding.

\subsection{Language and Visual Understanding}

It may seem counterintuitive that without visual inputs, language alone could come close to understanding images. However, as mentioned earlier, congenitally blind individuals demonstrate a similar understanding of many visual concepts to sighted individuals \citep{kim2021shared,kerr1983role, aleman2001visual, zimler1983imagery, connolly2007effect}. This suggests that there is potential to use linguistic descriptions to build accurate representations of visual input.

Additionally, recent work suggests that purely text-based LLMs create representations that correspond well with those of humans in settings that range from sensory domains such as taste, color, and timbre \citep{marjieh2023language}, image/video/audio representations \citep{marjieh2022predicting, marjieh2022words}, and even abstract concepts \citep{xu2023does}. These findings suggest that text is sufficient to give LLMs an understanding of the basic sensory inputs, visual similarity, and abstract concept associations that underlie our representation of the world. Based on these results, we hypothesize that language may play a substantial role in visual understanding; we extend these works by moving beyond representational similarity and into visual understanding to explore this hypothesis.

\section{Exploring the Space of Cognitive Architectures}

The question of what roles language and vision play in learning from limited data about the visual world around us can be framed as a question about {\em cognitive architectures}.
Cognitive architectures provide both a theoretical structure for understanding the components of complex cognitive systems and a practical framework for implementing computational models. First introduced in \citet{newell1972human}, they have since been used as the basis for models of human cognition, such as in \citet{anderson2013architecture}. More recently, researchers have looked to apply cognitive architectures to systems based upon LLMs \citep{sumers2023cognitive,sun2024can} as a way to better understand, design, and compare the assumptions in these systems. 

\begin{figure}[hbt!]
    \centering
    \includegraphics[width=0.85\linewidth,trim={0 0.05in 0 0.08in},clip]{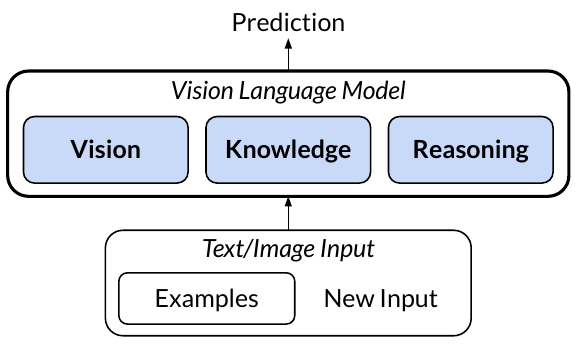}
        \vspace{-0.12in}
        \caption{Hypothesized cognitive architecture for intelligent visual recognition systems. A full architecture consists of visual processing, prior knowledge, and reasoning components and is given relevant training examples prior to testing.
        } 
\label{fig:cog_arch}
\end{figure}

We can use the cognitive architecture approach to hypothesize the \textit{abstract} components that contribute to learning to solve a new visual recognition problem. \figref{fig:cog_arch} shows a simple architecture for accomplishing this. It identifies four important components of learning visual understanding from limited data, three of which are internal to the model: \textbf{visual processing}, prior \textbf{knowledge}, and \textbf{reasoning}. The fourth is relevant training \textbf{examples}. Both humans and \textbf{V}LMs can be analyzed in terms of this architecture. In \textbf{V}LMs, the image model provides visual processing, knowledge is the stored internal representations of both the vision and language models, and reasoning is the non-trivial method of piecing information together to form a final prediction.

 Having described \textbf{V}LMs and LLMs in these terms, we can then effectively ablate specific components of these models to determine their significance. Concretely, we differentiate between LLMs and \textbf{V}LMs depending on the input type: text only for LLMs and text and images for \textbf{V}LMs.  

 \begin{figure*}[hbt!]
    \centering
    \includegraphics[width=0.95\linewidth]{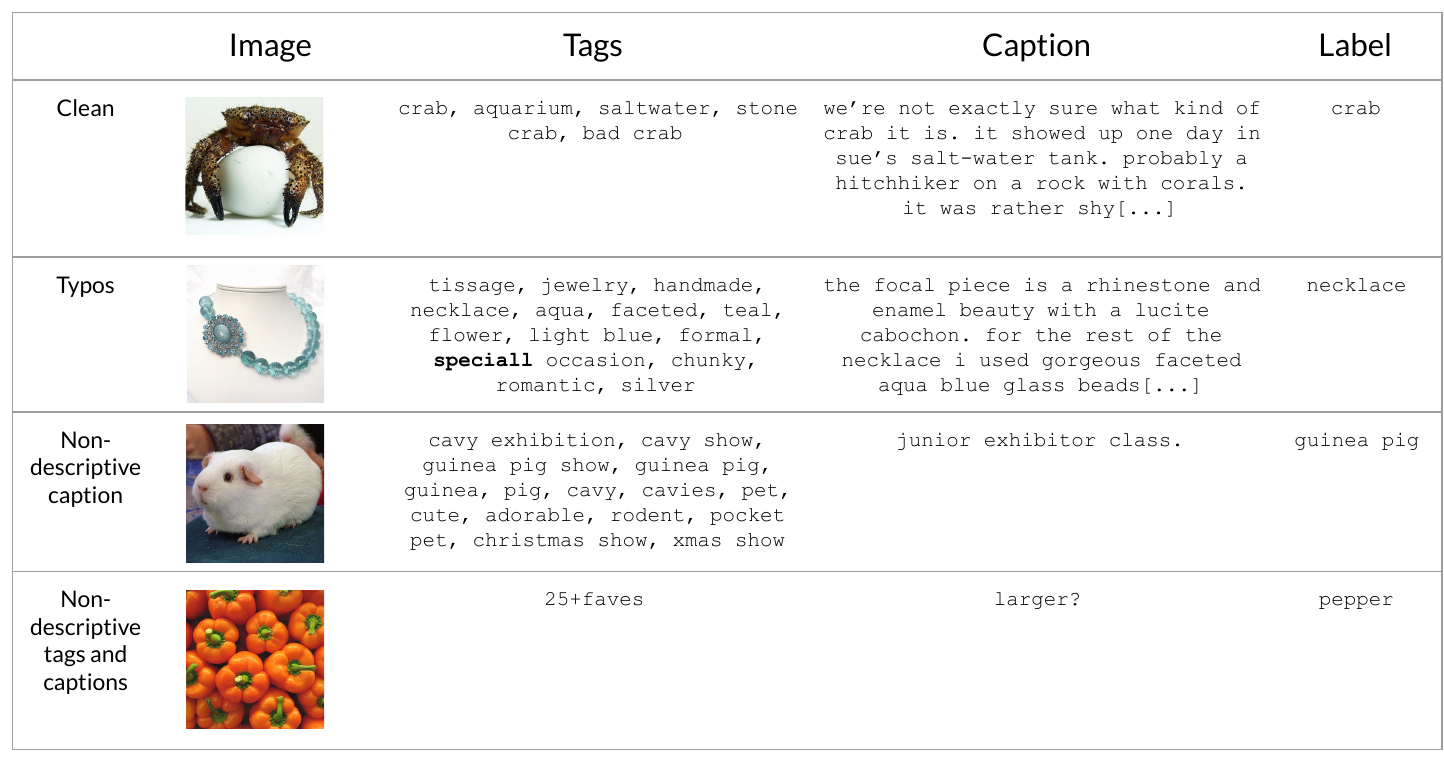}
    \vspace{-0.2in}
        \caption{
            Examples of images, corresponding tags, captions, and the class labels. \texttt{[...]} indicates the text was truncated due to length. First row represents a clean example and the bottom three demonstrate variance in the data.
        } 
    \vspace{-0.15in}
\label{fig:data_examples}
\end{figure*}

 We use the task of image recognition to measure visual understanding from limited data. We begin with a full \textbf{V}LM containing all three components of the architecture (i.e., ``Vision + Knowledge + Reasoning"). To study language's abilities to enable visual understanding, we isolate the LLM (abstaining from visual processing), use semantically meaningful text representations of each image (rather than the pixel-based image), and ablate components of the model. We also remove language from the \textbf{V}LM by utilizing a vision-only model with images and perform similar ablations. Removing each component and re-testing the performance of the system illustrates the contribution and necessity of each component.

Our simulations address three specific research questions. First, how close can an LLM (lacking visual processing) get to the performance of a \textbf{V}LM (with visual processing)?
Second, how do the LLM's various components contribute to its performance? And third, how does a vision model lacking language compare to the \textbf{V}LM?

\section{Methods}

We focus on the task of image recognition to measure visual understanding. In this section, we explain our dataset curation process, describe our instantiations of each cognitive architecture, and outline the three simulations we conducted. 

\subsection{Data Curation}
We sought a large dataset with semantically meaningful text-image pairs with associated class labels from a finite set of classes. We chose ImageNet Captions \citep{fang2022data}, which is based on the ImageNet Large Scale Visual Recognition Challenge 2012 \citep{russakovsky2015imagenet} training set. Each image is associated with an ImageNet-1K class label as well as a title, caption, and/or tags from Flickr, where the images were originally uploaded. As visualized in \figref{fig:data_examples}, some of the tags and captions are non-informative. However, this variability enables us to probe the limits of accurate classification while facing limits on descriptiveness due to realistic noise. Further, we used baselines and compared results across consistent test sets such that the effect of uninformative text is equivalent across conditions. 
For our simulations, we only used the tags as text representations due to their high information density, and we generally found that they are sufficiently descriptive of the images to make classification possible.

Preprocessing included cleaning and filtering the data and selecting a subset to use for classification. The ImageNet Captions dataset originally contained 463.6K images spanning 999 of the ImageNet-1K classes with between 7 to 961 samples per class. Text preprocessing entailed replacing all continuous white space with a single space, removing HTML tags, and converting all text to lowercase. Next, we only kept samples with all three text modalities (titles, captions, and image tags) and that were identified as English by the Python \texttt{langid} package. Lastly, due to large differences in the number of samples per class and the infeasibility of using all 999 classes, we selected 103 of the ImageNet-1K classes and re-labeled the images into 86 basic-level class categories based on semantic relatedness, roughly balancing the number of images per category. If a sample's text contained its new label, we replaced the occurrence with the text \texttt{[OMIT]}. After imposing new class labels, we split the images for each class into training, validation, and test pools using a 50/10/40 split. In our simulations, we selected a few training examples for each class from the training pool, reserved the validation pool for hyperparameter tuning (e.g., prompt wording), and reported results on a subset of the test pool due to cost limitations.

\begin{figure*}[hbt!]
    \centering
    \includegraphics[width=1.0\linewidth,trim={0 0.05in 0 0.13in},clip]{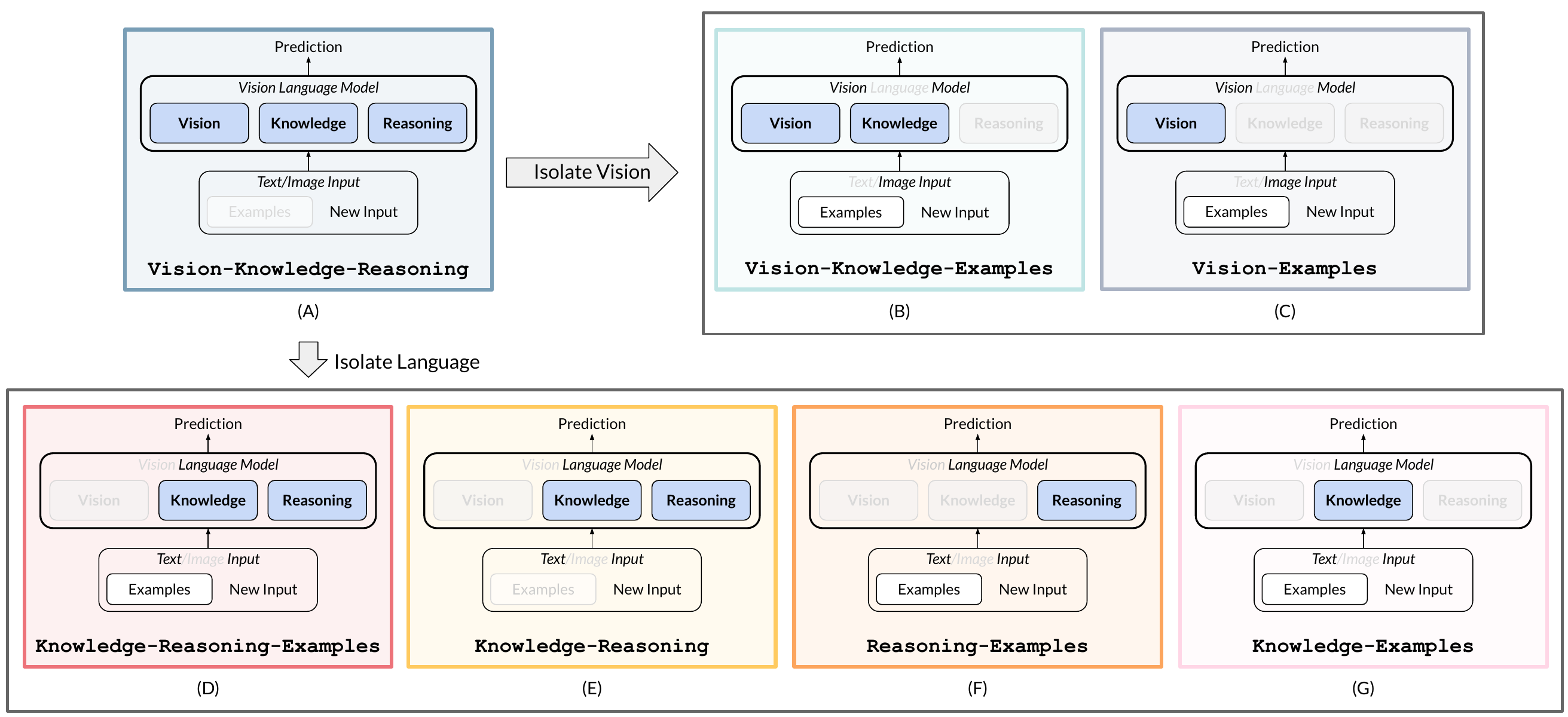}
        \vspace{-.2in}
        \caption{Cognitive architectures used in simulations. \textbf{Top Left}: A full vision language model but lacking examples. \textbf{Top Right}: Possible architectures when we remove language. \textbf{Bottom}: Possible architectures when removing vision. The bottom left box shows a full LLM, and the right three represent removing one component at a time: examples, knowledge, or reasoning. Colored boxes are used to differentiate these models in subsequent presentation of the results.} 
        \vspace{-0.1in}
\label{fig:cog_arch_all}
\end{figure*}

\subsection{Implementations of Cognitive Architectures}

Here, we provide implementation details of how we ablated various components to instantiate the cognitive architectures found in \figref{fig:cog_arch_all}.

\noindent\textbf{Vision-Knowledge-Reasoning (\figref{fig:cog_arch_all}A).} We originally attempted to implement the ideal architecture in \figref{fig:cog_arch} using GPT-4V but encountered limits on the number of images per prompt using our API, making this infeasible. Instead, we removed examples and implemented the cognitive architecture in \figref{fig:cog_arch_all}A by querying a \textbf{V}LM, GPT-4V, to classify images without providing any examples prior to testing.

\noindent\textbf{Knowledge-Reasoning-Examples (\figref{fig:cog_arch_all}D).} We then removed visual processing from the \textbf{V}LM, shown by \figref{fig:cog_arch_all}D, by using its language backbone: GPT-4\footnote{The model can only receive text input, so we consider it an LLM. However, GPT-4 may have been trained on both images and text. Additionally, GPT-4V technically uses GPT-4-Turbo, a smaller variant of GPT-4, as a backbone. However, preliminary experiments comparing the two showed similar performance.}. To use all components, including examples, we queried GPT-4 by providing 3 text examples per class with their corresponding labels prior to testing, using the tags associated with each image as semantic text representations of the image. The next three architectures each ablate one component from this LLM.

\noindent\textbf{Knowledge-Reasoning (\figref{fig:cog_arch_all}E).} To ablate examples, we simply removed examples from the prompt for GPT-4. This is similar to \texttt{Vision-Knowledge-Reasoning} (\figref{fig:cog_arch_all}A), but with tags instead of images. By removing training examples, we prohibited the model from sampling the data distribution.

\noindent\textbf{Reasoning-Examples (\figref{fig:cog_arch_all}F).} Knowledge cannot be cleanly removed due to its distributed nature in LLMs, so instead we limited its usefulness. To do so, we provided GPT-4 with 3 examples per class before test time, similar to \texttt{Knowledge-Reasoning-Examples} (\figref{fig:cog_arch_all}D), but we use fake (i.e. non-English) words with a 1:1 mapping to the true class labels. Because the fake word labels bear no semantic significance and break the semantic connection between the tags and labels, this setup \textit{diminishes} the utility of prior knowledge. The pool of fake words was generated by GPT

\noindent\textbf{Knowledge-Examples (\figref{fig:cog_arch_all}G).} Because reasoning cannot be entirely removed, we reduced it to be trivial--where our threshold for non-trivial reasoning is adapting to new tasks without adding or modifying weights. To do so, we replaced GPT-4 with OpenAI's \texttt{text-embedding-ada-002} \citep{openaiImprovedEmbedding} (henceforth referred to as Ada) embedding model with a linear classifier. We used Ada to obtain embeddings of each sample's tags, then trained a linear classifier with 3 examples per class. Because the Ada-based linear classifier must learn parameters to perform classification, while GPT does not, this model has trivial reasoning capabilities at best.

\begin{figure*}[hbt!]
    \centering
    \includegraphics[width=0.9\linewidth,trim={0.10in 0.35in 0.1in, 0.15in},clip]{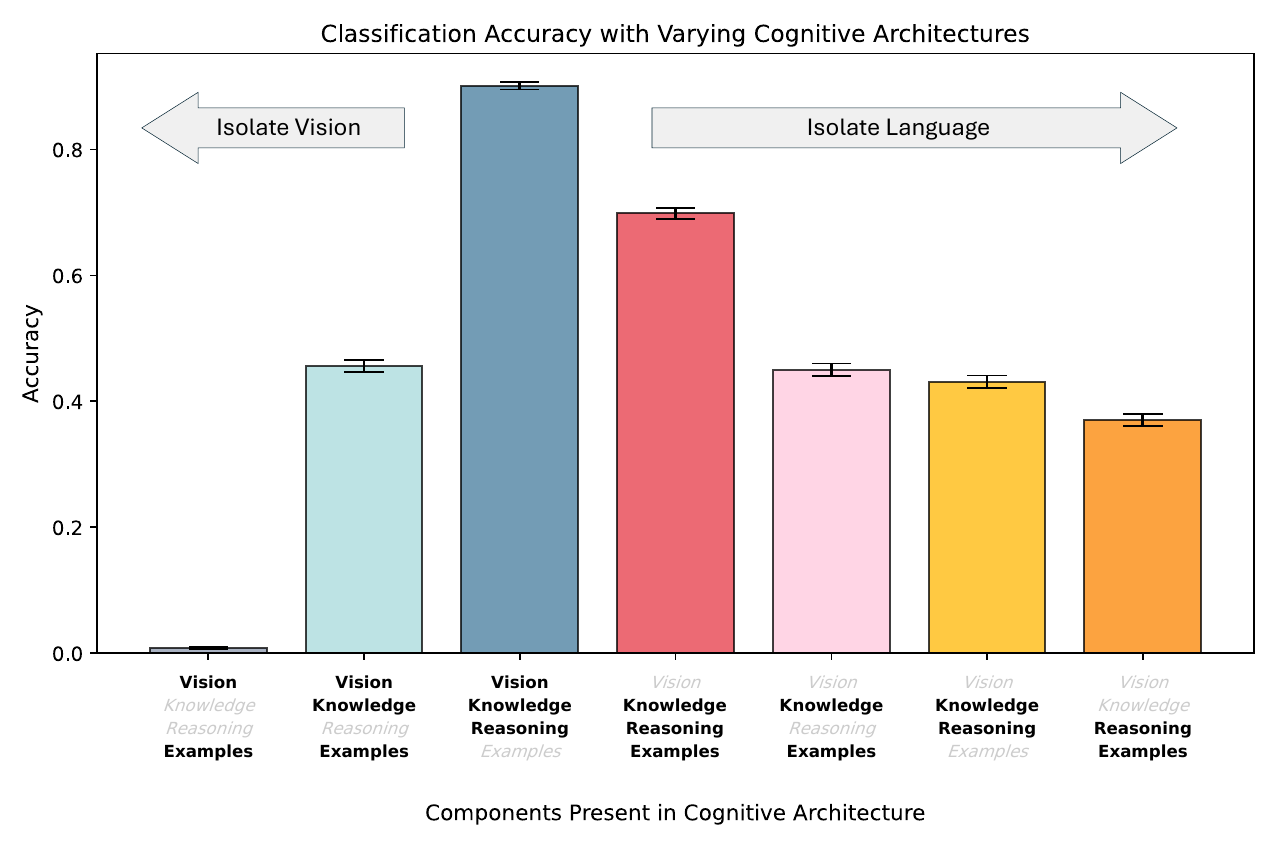}
        \vspace{-0.2in}
        \caption{
            Results from simulations. The horizontal axis lists components present in each setting where \textbf{bold} indicates component present and \textcolor{gray}{\textit{gray and italicized}} indicates component missing. Vision, knowledge, and reasoning are part of a model's architecture and examples refers to the model seeing relevant examples of the new task. The highest bar in dark blue titled \texttt{Vision-Knowledge-Reasoning} is the full \textbf{V}LM. To the right we isolated language and its various components, and to the left, we isolated components of vision. Bar colors correspond to the boxes around models shown in \figref{fig:cog_arch_all}. Error bars represent standard error of the mean calculated by treating each trial as a binomial random variable.
        } 
        \vspace{-0.1in}
\label{fig:pull_fig}
\end{figure*}

\noindent\textbf{Vision-Knowledge-Examples (\figref{fig:cog_arch_all}B).} Now, we outline the implementations of cognitive architectures where vision is isolated. However, because OpenAI did not release the vision model of GPT-4V, we used a ResNet50 \citep{he2016deep} classifier as a proxy. The model was pre-trained on the PASS image dataset \citep{asano21pass}, equipping it with knowledge. We recognize that the scale of the data and model is much smaller than that of GPT-4V, but we leave this limitation for future work. Similar to \texttt{Knowledge-Examples} (\figref{fig:cog_arch_all}G), we used the pre-trained model to obtain embeddings, now from images, then trained a linear classifier using 3 images per class. The ResNet50 classifier lacks reasoning for the same reason as the Ada classifier.

\noindent\textbf{Vision-Examples (\figref{fig:cog_arch_all}C).} We removed prior knowledge from the vision-only architecture by randomly initializing our ResNet50 classifier (i.e., no pre-training). We used the 3 samples per class to train the full model from scratch, effectively removing information beyond the small number of examples.

\noindent\textbf{Additional Implementation Details.} Prompts for models with reasoning (i.e., GPT-4V and GPT-4) contained the following: task instructions, the list of classes, shuffled training examples with our class labels (if providing examples), and a new sample.
We accessed language models using the Azure OpenAI Python API with the temperature set to $0$ to encourage consistency across responses. When the prompt length was too long for GPT-4's 8K token limit, we used GPT-4-32K with a 32K token limit. Linear classifiers were implemented using the \texttt{sklearn.neural\_network.MLPClassifier} package in Python with no hidden layer, and we performed a hyperparameter search to determine the learning rate. The pre-trained ResNet50 was obtained from the Pytorch model repository. Models were evaluated on a test set of $n=2500$ samples using the 86 basic-level class categories we defined.

\subsection{Simulations}

We conducted simulations to address three questions: 

\textbf{Simulation 1:} \textbf{How close can an LLM (lacking visual processing) get to the performance of a \textbf{V}LM (with visual processing)?} We compared the full \texttt{Vision-Knowledge-Reasoning} (\figref{fig:cog_arch_all}A) \textbf{V}LM to the full \texttt{Knowledge-Reasoning-Examples} (\figref{fig:cog_arch_all}D) LLM. The goal was to obtain a lower bound on the value of language in visual understanding by removing visual processing.

\textbf{Simulation 2:} \textbf{How do the LLM's various components contribute to its performance?} Here, we used the full \texttt{Knowledge-Reasoning-Examples} (\figref{fig:cog_arch_all}D) LLM as a baseline and compared it to the following three cognitive architectures, each removing one of the components: \texttt{Knowledge-Reasoning} (\figref{fig:cog_arch_all}E), \texttt{Reasoning-Examples} (\figref{fig:cog_arch_all}F), and \texttt{Knowledge-Examples} (\figref{fig:cog_arch_all}G). By removing one component at a time, we could evaluate how necessary each is to understanding in the absence of visual processing.

\textbf{Simulation 3:} \textbf{How does a vision model lacking language compare to the VLM?} Lastly, we compared the vision-only architectures, \texttt{Vision-Knowledge-Examples} (\figref{fig:cog_arch_all}B) and \texttt{Vision-Examples} (\figref{fig:cog_arch_all}C) to the full \textbf{V}LM \texttt{Vision-Knowledge-Reasoning} (\figref{fig:cog_arch_all}A). The purpose was to understand how visual processing in the absence of language could perform, helping us identify a lower bound of both the necessity of language in visual understanding and of knowledge when lacking language.

\section{Results}

Results from all simulations can be found in \figref{fig:pull_fig}. 

\textbf{Simulation 1:} Here, we compared a \textbf{V}LM without examples to a full LLM with examples. In \figref{fig:pull_fig}, the \textbf{V}LM performance is the dark blue bar labeled \texttt{Vision-Knowledge-Reasoning} and the full LLM is the red bar in the center labeled \texttt{Knowledge-Reasoning-Examples}. The \textbf{V}LM baseline obtains an accuracy of 90.1\% (Standard Error of the Mean (SEM)=$0.60\%$) while the LLM only reaches 69.84\% (SEM=$0.92\%$, $p < 0.001$), maintaining just over 75\% of the \textbf{V}LM's performance.

\textbf{Simulation 2:} We ablated the language components of a full LLM, demonstrated by moving to the right in \figref{fig:pull_fig}. P-values are reported compared to the full LLM from Simulation 1. As shown by the orange bar labeled \texttt{Reasoning-Examples} on the far right, when limiting access to prior knowledge by replacing true labels with fake words, the language model performance drops from 69.84\% to 37.00\% (SEM=$0.97\%$, $p < 0.001$). When eliminating training examples, (see the yellow bar labeled \texttt{Knowledge-Reasoning}), model performance drops to 43.10\% (SEM=$0.99\%$, $p < 0.001$). Lastly, when reducing non-trivial reasoning from the language model, performance drops to 45.00\% (SEM=$0.99\%$, $p < 0.001$), demonstrated by the pink bar labeled \texttt{Knowledge-Examples}. 

\textbf{Simulation 3:} We removed language entirely, demonstrated by the bars on the left side of the \textbf{V}LM performance. We report p-values compared to the \textbf{V}LM performance. When the vision-only model has prior knowledge, performance drops to 45.60\% (SEM=$1.00\%$, $p < 0.001$) as seen by the light blue bar labeled \texttt{Vision-Knowledge-Examples} in \figref{fig:pull_fig}. However, when prior knowledge is removed, the model only achieves 0.84\% (SEM=$0.18\%$, $p < 0.001$) --approximately random chance. This is seen by the gray bar labeled \texttt{Vision-Examples} on the far left.

\section{Discussion}
Our work aims to investigate the contribution of aspects of language and visual processing to understanding the visual world. Based on our results, visual processing is a large component of visual intelligence, but sophisticated LLMs--when equipped with strong prior knowledge, reasoning mechanisms, and a few training examples--perform surprisingly well. However, only when all three are present does the model without visual processing perform remotely close to a joint vision and language system. When removing any one of these three, performance drops significantly and by similar amounts compared to removing any other component. These observations suggest that all three--knowledge, reasoning, and examples--are necessary and have approximately equal importance for visual understanding via textual inputs.

On the vision side, a vision model with knowledge but no reasoning achieves approximately half the performance of the \textbf{V}LM, which is comparable to a language model missing any one component. However, removing prior knowledge reduces performance to essentially random chance. This suggests that, when lacking sophisticated reasoning, sufficient prior knowledge is necessary for visual understanding.

\subsection{Limitations and Future Work}
While the results of our simulations provide insight into the roles of language and vision in learning from limited data, we acknowledge a few limitations. First, because the text we use as our language input was obtained from the image's source rather than annotated, they vary in descriptiveness. Future work should explore collecting new tags via crowdsourcing processes like STEP-Tag \citep{marjieh2022words} or by using a \textbf{V}LM. We also acknowledge the limitations of solely running our evaluations on GPT due to the closed nature of its architecture and training data. To increase the robustness of our findings, we aim to repeat experiments on open-source models such as LLaMA \citep{touvron2023llama} and LLaVA \citep{liu2023llava}. 
Lastly, as mentioned earlier, we acknowledge the pre-trained image model is smaller than the \textbf{V}LM and uses a much smaller dataset, which may be limiting its performance (and hence underestimating the contribution of prior knowledge in the vision case). We hope to find or train image models on comparable amounts of data to further investigate our hypotheses.

\subsection{Conclusion} 

New AI models bring with them the opportunity to understand the factors that give rise to complex and intelligent behavior. Here, we used \textbf{V}LMs to analyze the contributions of visual processing and language to visual understanding and further dissected language into more granular components. Our findings show that a LLM with prior knowledge, reasoning, and a few relevant examples can surprisingly recover three-quarters of a powerful \textbf{V}LM's performance, despite not having access to the actual images. Further, the synergy of all three language components mentioned above is necessary to achieve this performance; removing any one severely hinders the model's abilities. Lastly, vision models lacking language fall behind full language models and do not significantly outperform language models missing one of the key mechanisms. This suggests that a language of thought \citep{fodor1975language} may indeed be valuable even in perceptual tasks like visual classification. While these findings do not directly shed light on how humans--a very different kind of system--use vision and language, they provide a lower bound on the signal that can be derived from these sources.

In addition to identifying language's role in visual understanding, this work also contributes to demystifying high-performing \textbf{V}LMs and LLMs by applying cognitive science approaches to understanding the powerful black boxes. 
While we are far from understanding \textbf{V}LMs and LLMs in their entirety, we believe that the approach we have developed here--specifying a cognitive architecture and exploring its structure through a series of ablations--is a powerful way to begin identifying the key components of these systems and ultimately increasing their transparency.

\subsubsection{Acknowledgements.} This material is based upon work supported by Microsoft Azure credits supplied to Princeton, by a Microsoft Foundation Models grant, and by the National Science Foundation under Grant No. 2107048 and 2145198. Any opinions, findings, and conclusions or recommendations expressed in this material are those of the author(s) and do not necessarily reflect the views of the National Science Foundation. We would also like to thank (in alphabetical order by last name) Jihoon Chung, Amaya Dharmasiri, Max Gonzalez Saez-Diez, Ryan Liu, Maya Malaviya, Shruthi Santhanam, Xindi Wu, and William Yang for their input and feedback.

\bibliographystyle{apacite}

\setlength{\bibleftmargin}{.125in}
\setlength{\bibindent}{-\bibleftmargin}

\bibliography{references}

\end{document}